\documentclass[11pt,a4paper]{article}
\usepackage[hyperref]{eacl2021}
\usepackage{times}
\usepackage{latexsym}

\usepackage{times}
\usepackage{url}
\usepackage{latexsym}
\usepackage{caption}
\usepackage{algorithm}
\usepackage{algorithmic}
\usepackage{soul}
\usepackage{multicol}
\usepackage{multirow}
\usepackage{booktabs}
\usepackage{url}
\usepackage{todonotes}
\usepackage{caption}
\usepackage{subcaption}
\usepackage{arydshln}
\usepackage{amssymb}
\usepackage{multicol}
\usepackage{amsmath}
\usepackage{booktabs}
\usepackage{bigdelim}
\usepackage{amsmath}
\usepackage{colortbl}
\usepackage{todonotes}
\usepackage{hyperref}
\usepackage{graphicx}
\usepackage{microtype}

\aclfinalcopy % Uncomment this line for the final submission
\title{Meta-Learning for Effective Multi-task and  Multilingual Modelling}

\author{
Ishan Tarunesh\textsuperscript{\rm 1} \hspace{0.4em} Sushil Khyalia\textsuperscript{\rm 1} \hspace{0.4em} 
 Vishwajeet Kumar\textsuperscript{\rm 2} \\ \bf{Ganesh Ramakrishnan}\textsuperscript{\rm 1}   \hspace{0.4em} \bf{Preethi Jyothi}\textsuperscript{\rm 1}  \\
$^1$ Indian Institute of Technology Bombay\\
$^2$ IBM India Research Lab \\
\texttt{\{ishan, sushil, ganesh, pjyothi\}@cse.iitb.ac.in} \\
\texttt{vishk024@in.ibm.com}
}

\date{}

\begin{document}
\maketitle
\begin{abstract}
Natural language processing (NLP) tasks (e.g.~question-answering in English) benefit from knowledge of other tasks ({\em e.g.}, named entity recognition in English) and knowledge of other languages ({\em e.g.}, question-answering in Spanish). Such shared representations are typically learned in isolation, either across tasks or across languages. In this work, we propose a meta-learning approach to learn the interactions between both tasks and languages. We also investigate the role of different sampling strategies used during meta-learning. We present experiments on five different tasks and six different languages from the XTREME multilingual benchmark dataset~\cite{pmlr-v119-hu20b}. Our meta-learned model clearly improves in performance compared to competitive baseline models that also include multi-task baselines. 
We also present zero-shot evaluations on unseen target languages to demonstrate the utility of our proposed model.
%\small
% Meta-Learning algorithms try to learn a general representation of data belonging to multiple tasks. In this work, we analyse an optimisation-based meta-learning algorithm Reptile in a multilingual multitask setup. Additionally, we examine how the performance of meta-learning changes by changing the sampling strategy. We use XTREME benchmark for evaluation and observe an improved performance on QA, NLI, and PAWS for all languages. 
% We propose to train a meta-learning model in multi task multi lingual setup.

% Xtreme dataset. Better sampling strategy. 
%  In any meta learning setup \vish{In meta learning setting,} given a distribution of tasks T with the aim of finding optimal $\theta^{*}$ so that the task-specific fine-tuning \vish{leads to better performance} is more effective.
\end{abstract}
\section{Introduction}
\label{introduction}

% With recent advances in deep learning, 
Multi-task and multilingual learning are both problems of long standing interest in natural language processing. Leveraging data from multiple tasks and/or additional languages to benefit a target task is of great appeal, especially when the target task has limited resources. When it comes to multiple tasks, it is well-known from prior work on multi-task learning~\cite{liu2019multi,kendall2018multi,multitask-with-attention,deep-multitask-representation-learning} %\ishan{Added this https://arxiv.org/abs/1803.10704 and https://arxiv.org/abs/1605.06391} 
that jointly learning a model across tasks can benefit the tasks mutually.  For multiple languages, the ability of deep learning models to learn effective embeddings has led to their use for joint learning of models across languages~\cite{conneau-etal-2020-unsupervised,lample2019cross,artetxe2019massively}; learning cross-lingual embeddings to aid languages in limited resource settings is of growing interest~\cite{kumaretal2019cross,wang2017multi,adams2017cross}.
%\ishan{Maybe these? https://www.aclweb.org/anthology/I17-2065/, https://arxiv.org/abs/1810.12836, https://www-nlp.stanford.edu/pubs/schuster2019crosslingual.pdf,https://www.aclweb.org/anthology/E17-1088.pdf} 
Let us say we had access to $M$ tasks across $N$ different languages - {\em c.f.} Table~\ref{tab::datasetdetails} that outlines such a matrix of tasks and languages from the XTREME benchmark~\cite{pmlr-v119-hu20b}. How do we perform effective joint learning across tasks and languages? Are there specific tasks or languages that need to be sampled more frequently for effective joint training? Can such sampling strategies be learned from the data? 

\begin{figure*}[!tb]
    \centering
    \includegraphics[width=0.8\textwidth]{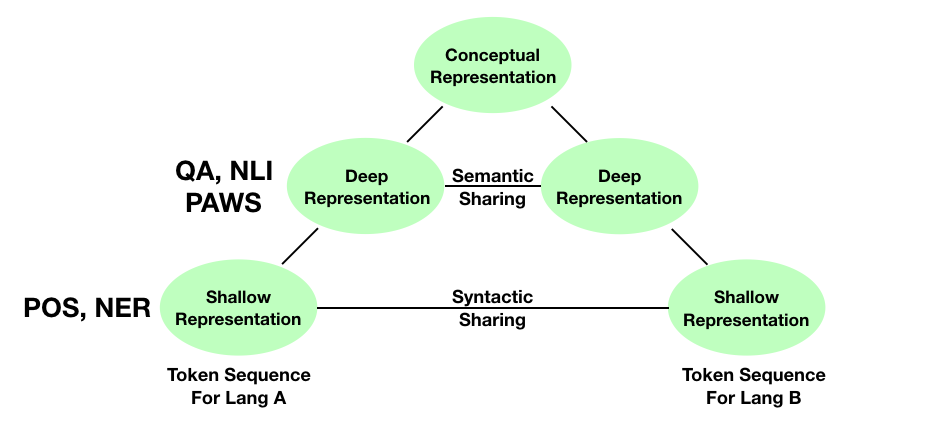}
    \caption{\small Illustration derived from Vauquois Triangle to linguistically motivate our setting. POS and NER being lower down in the representations (and are thus `shallower') are further away from the same task in another language. QA, XNLI and PAWS being higher up in the representations (and are thus `deeper') are closer to the same task in another language.}
    \label{fig:vtriangle}
\end{figure*}

In this work, we adopt a meta-learning approach for efficiently learning parameters in a shared parameter space across multiple tasks and multiple languages. Our chosen tasks are question answering (QA), natural language inference (NLI), paraphrase identification (PA), part-of-speech tagging (POS) and named entity recognition (NER). The tasks were chosen to enable us to employ a gamut of different types of language representations needed to tackle problems in NLP. In Figure~\ref{fig:vtriangle}, we illustrate the different types of representations by drawing inspiration from the Vauquois Triangle~\cite{Vauquois1968ASO}, well-known for machine translation, and situating our chosen tasks within such a triangle. Here we see that POS and NER are relatively `shallower' analysis tasks that are token-centric, while QA, NLI and PA are `deeper' analysis tasks that would require deeper semantic representations. This representation suggests a strategy for effective parameter sharing. For the deeper tasks, the same task in different languages could have representations that are closer and hence benefit each other, while for the shallower tasks, keeping the language unchanged and exploring different tasks might be more beneficial. Interestingly, this is exactly what we find with our meta-learned model and is borne out in our experimental results. We also find that as the model progressively learns, the meta-learning based models for the tasks requiring deeper semantic analysis benefit more from joint learning compared to the shallower tasks.

%In this work, we adopt a meta-learning approach for efficiently learning parameters in a shared parameter space across multiple tasks and multiple languages. Our chosen tasks are question answering (QA), cross-lingual natural language inference (XNLI), cross-lingual paraphrase identification (XPA)\pj{Either call it xpa or expand PAWS} \ishan{It is not exactly cross-lingual in the sense we are not using dataset with cross-lingual sentences - for both XNLI and XPA}, part-of-speech tagging (POS) and named entity recognition (NER). For each of these tasks, we use data from multiple languages. We note here that these tasks span the gamut of different types of language reasoning that appear in the Vauquois Triangle~\cite{Vauquois}, illustrated in Figure~\ref{fig:vtriangle}. POS and NER are relatively `shallower' analysis tasks that are token-centric, while QA, XNLI and PAWS are `deeper' analysis tasks that would require deeper semantic representations. The triangle suggests that the shallower tasks may not benefit as much as the deeper tasks across languages. We see this borne out in our empirical analysis as well; as the proposed algorithm progressively learns, the meta-learning based models for the tasks requiring deeper semantic analysis benefit more from joint learning across languages compared to the shallower tasks. \todo{SS: Add the vtriangle figure/citations.}

With access to multiple tasks and languages during training, the question of how to sample effectively from different tasks and languages also becomes important to consider. We investigate different sampling strategies, including a parameterized sampling strategy, to assess the influence of sampling across tasks and languages on our meta-learned model. 

Our main contributions in this work are three-fold:

\begin{itemize}
\item We present a meta-learning approach that enables effective sharing of parameters across multiple tasks and multiple languages. This is the first work, to our knowledge, to explore the interplay between multiple tasks at different levels of abstraction and multiple languages using meta-learning. We show results on the recently-released XTREME benchmark and observe consistent improvements across different tasks and languages using our model. We also offer rules of thumb for effective meta-learning that could hold in larger settings involving additional tasks and languages. 
\item We investigate different sampling strategies that can be incorporated within our meta-learning approach and examine their benefits.
\item We evaluate our meta-learned model in zero-shot settings for every task on target languages that never appear during training and show its superiority compared to competitive zero-shot baselines.
\end{itemize}

\section{Related Work}
\label{related}
We summarize three threads of related research that look at the transferability in models across different tasks and different languages: multi-task learning, meta-learning and data sampling strategies for both multi-task learning and meta-learning. Multi-task learning~\cite{Caruana1993MultitaskLA} has proven to be highly effective for transfer learning in a variety of NLP applications such as question answering, neural machine translation,  {\em etc.}~\cite{mccann2018natural,hashimotoetal2017joint,Chen2018MetaML,Kiperwasser2018ScheduledML}. Some multi-task learning approaches~\cite{jawanpuriajmlr15} have attempted to identify clusters (or groups) of related tasks based on end-to-end convex optimization formulations. 
%A second thread of work is based on the idea of meta-learning which aims to learn a better initialization for the model to quickly adapt to different albeit related tasks.
Meta-learning algorithms~\cite{Nichol2018OnFM} are highly effective for fast adaptation and have recently been shown to be beneficial for several machine learning tasks~\cite{Santoro2016MetaLearningWM,pmlr-v70-finn17a}. %There has been growing interest in recent years in using meta-learning for NLP. 
\newcite{gu2018meta} use a meta-learning algorithm for machine translation to leverage information from high-resource languages. \newcite{dou2019investigating} investigate multiple model agnostic meta-learning algorithms for low-resource natural language understanding on the GLUE~\cite{wang2018glue} benchmark.

Data sampling strategies for multi-task learning and meta-learning form the third thread of related work. A good sampling strategy has to account for the imbalance in dataset sizes across tasks/languages and the similarity between tasks/languages. A simple heuristic-based solution to address the issue of data imbalance is to assign more weight to low-resource tasks or languages~\cite{aharoni2019massively}.
\newcite{Arivazhagan2019MassivelyMN} define a temperature parameter which controls how often one samples from low-resource tasks/languages. The MultiDDS algorithm, proposed by \newcite{Wang2020BalancingTF}, actively learns a different set of parameters for sampling batches given a set of tasks such that the performance on a held-out  set is maximized.
We use a variant of MultiDDS as a sampling strategy in our meta-learned model.
\newcite{nooralahzadeh2020zero} is most similar in spirit to our work in that they study a cross-lingual and cross-task meta-learning architecture but only focus on zero-shot and few-shot transfer for two natural language understanding tasks, NLI and QA. In contrast, we study many tasks in many languages, in conjunction with sampling strategies, and offer concrete insights on how best to guide the meta-learning process when multiple tasks are in the picture.

%for two natural language understanding task with focus on zero-shot and few-shot cross-lingual transfer. Previous research focus mainly on meta-transfer either across tasks or across languages, whereas we study meta transfer across task-language pairs. 

%\ishan {General Suggestions - I think we should limit ourself to these 4 papers
%\begin{itemize}
%	\item {Meta-Learning for Low-Resource Neural Machine Translation}
%	\item {Investigating Meta-Learning Algorithms for Low-Resource Natural Language Understanding Tasks}
%	\item {Balancing Training for Multilingual Neural Machine Translation}
%	\item {Scheduled Multi-Task Learning: From Syntax to Translation (Not sure about this also)}
%\end{itemize}
%}
% ===============

% In this paper, we propose and study meta learning algorithm in multi-task multi-lingual setup leveraging both multiple tasks and multiple languages.

\section{Methodology}
Our setting is pivoted on
%characterized by 
a \textbf{grid} of tasks and languages (with some missing entries as shown in Table~\ref{tab::datasetdetails}).
% We can have the same  task to be solved in multiple languages and we could have multiple tasks in a single language. We visualize our dataset as a grid, with some missing entries.. 
Each row of the grid corresponds to a single \textbf{task}. A cell of the grid corresponds to a Task-Language pair which we refer to as a TL pair (\textbf{TLP}).
We denote by $q_{i} = |\mathcal{D}_{train}^{i}|/\big(\sum_{k = 1}^{n} |\mathcal{D}_{train}^{k}|\big)$, the fraction of the dataset size for the $i^{th}$ TLP and by $P_{\mathcal{D}}(i)$, the probability of sampling a batch from the $i^{th}$ TLP during meta training. The distribution over all TLPs, {\em viz.},  is a Multinomial (say $\mathcal{M}$) over $P_{\mathcal{D}}(i)$s.

\subsection{Our Meta-learning Approach}

The goal in the standard meta learning setting is to obtain a model that generalizes well to new test/target tasks given some distribution over training tasks. This can be achieved using optimization-based meta-learning algorithms that modify the learning procedure in order to learn a good initialization of the parameters. This  can serve as a useful starting point that can be further fine-tuned on various tasks. \newcite{pmlr-v70-finn17a} proposed a general optimization algorithm called Model Agnostic Meta Learning (MAML) that can be trained using gradient descent. MAML aims to minimize the following objective%~\ref{eqn:maml}
\begin{equation}
\label{eqn:maml}
    \min_{\theta} \sum_{T_{i} \sim \mathcal{M}} \mathcal{L}_{i} \left( U^{k}_{i} (\theta)\right)
\end{equation}
% $$\min_{\theta} \sum_{T_{i} \sim \mathcal{M}} \mathcal{L}_{i} \left( U^{k}_{i} (\theta)\right)$$
where $\mathcal{M}$ is the Multinomial distribution over TLPs, $\mathcal{L}_{i}$ is the loss and $U^{k}_{i}$ a function that returns $\theta$ after k gradient updates both calculated on batches sampled from $T_{i}$. Minimizing this objective using first order methods involves computing gradients of the form $\frac{\partial}{\partial \theta} U^{k}_{i} (\theta)$, leading to the expensive computation of second order derivatives. 
%First order MAML (FOMAML) approximates MAML by omitting the second order derivates.  
\newcite{Nichol2018OnFM} proposed an alternative first-order meta-learning algorithm named ``Reptile" with simple update rule:
\begin{equation}
\label{eqn:reptile}
\theta \leftarrow \theta + \beta \frac{1}{|\{T_{i}\}|} \sum_{T_{i} \sim \mathcal{M}} (\theta^{(k)}_{i} - \theta)
\end{equation}
% $$\theta = \theta + \beta \frac{1}{|T_{i}|} \sum_{T_{i} \sim \mathcal{M}} (\theta^{(k)}_{i} - \theta)$$
where $\theta_i^{(k)}$ is $U^{k}_{i} (\theta)$.
%(model parameters after taking $k$ gradient steps on batches of $T_{i}$ starting from the initial $\theta$. 
Despite its simplicity, a recent study by \newcite{dou2019investigating} showed that Reptile is atleast as effective as MAML in terms of performance. We therefore employed Reptile for meta learning in all our experiments. 

%\noindent
%Multi-task learning also has multiple tasks with the aim to maximise performance on a particular task by leveraging information from other task and preventing overfitting to one task. In the essence both Multi-tasks learning and meta learning are similar but they achieve their goal differently. Multi-task learning designs a curriculum of scaling losses of different tasks to jointly learn as well as regularise. \\
%\\
%\noindent
%Most of the work on Meta Learning is aimed towards zero shot or low resource setup in a very related or limited set of tasks. The two most important design choice in meta learning is the choice of auxiliary tasks and the process of sampling of these tasks in meta training.
%This work is aimed at analysing different sampling strategies in meta learning framework with a diverse setup of NLP tasks - Question Answering (QA), Natural Language Inference (NLI), Part Of Speech Tagging (POS), Named Entity Recognition (NER) and Paraphrase Identification (PAWS) and multiple languages per task. We also investigate how the meta training could be made target task aware by creating meaningful subsets of auxiliary tasks and updating sampling probabilities using feedbacks from development datasets using REINFORCE \cite{williams1992simple} based updates.

\begin{algorithm}[!htb]
\hspace*{\algorithmicindent} \textbf{Input:} ${\mathcal{D}}_{train}$ set of TLPs for meta training (Also ${\mathcal{D}}_{dev}$ for parametrised sampling) \\
\hspace*{\algorithmicindent} \hspace*{\algorithmicindent}\hspace*{\algorithmicindent}\hspace*{\algorithmicindent} Sampling Strategy (Temperature / MultiDDS) \\
\hspace*{\algorithmicindent} \textbf{Output:} The converged multi-task multilingual model parameters $\theta^{*}$
\caption{Our Meta-learning Approach}
\label{algo::reptile}
% \begin{multicols}{2}
\begin{algorithmic}[1]
    \STATE \textbf{Initialize} $P_{D}(i)$ depending on the sampling strategy
    \WHILE{not converged}
    \STATE $\triangleright$ \textit{Perform Reptile Updates}
    \STATE Sample $m$ TLPs $T_{1}, T_{2}, \dots, T_{m}$ from $\mathcal{M}$ \\
        \FOR{i = 1,2,\dots,m}
            \STATE $\theta_i^{(k)} \leftarrow U_{i}^{k}(\theta),$ denoting $k$ gradient updates  from $\theta$ on batches of TLP $T_{i}$ \label{algo:gradupdate}
        \ENDFOR
        \STATE $\theta \leftarrow \theta + \frac{\beta}{m}\sum_{i=1}^{m}(\theta_i^{(k)} - \theta)$ \label{algo:reptileupdate}
        %\STATE $\triangleright$ \textit{Load training data with $\psi$}
        %\STATE $X,Y\ \leftarrow\ \Phi$
        %\WHILE{$|X,Y|<M$}
        %   \STATE $\Tilde{i} \sim P_{D}(i,\psi)$
        %    \STATE $(x,y) \sim D_{train}^{\Tilde{i}}$
        %    \STATE $X,Y \leftarrow X,Y \cup x,y$
        %\ENDWHILE
        %\STATE $\triangleright$ \textit{Train the model for multiple steps}
        %\FOR{$x,y$ in $X,Y$}
        %    \STATE $\theta \leftarrow$ GradientUpdate$(\theta,\nabla_{\theta}l(x,y;\theta))$
        %\ENDFOR
        \IF{Sampling Strategy $\leftarrow$ MultiDDS}
             \FOR{$\mathcal{D}_{train}^{i}$ $\in$ ${\mathcal{D}}_{train}$}
            %     \STATE $x',y' \sim D_{train}^{i}$
            %     \STATE $g_{train} \leftarrow \nabla_{\theta}l(x',y';\theta))$
            %     \STATE $\theta' \leftarrow$ GradientUpdate$(\theta,g_{train})$
            %     \STATE $g_{dev} \leftarrow 0$
            %     \FOR{$D_{dev}^{j}$ $\in$ $\{{D}_{dev}\}$}
            %         \STATE $x_d,y_d \sim D_{dev}^{j}$
            %         \STATE $g_{dev} \leftarrow g_{dev}+\nabla_{\theta'}l(x_d,y_d;\theta'))$
            %     \ENDFOR
                \STATE $R(i;\theta) \leftarrow cos(g_{dev},g_{train})$,\ $g_{dev}$ is gradient on $\{\mathcal{D}_{dev}\}$ and $g_{train}$ is gradient on $\mathcal{D}_{train}^{i}$
            \ENDFOR
            \STATE $\triangleright$ \textit{Update Sampling Probabilities}
            \STATE $d_\psi \leftarrow \sum_{i=1}^{n}R(i;\theta)\cdot\nabla_\psi log(P_\mathcal{D}(i;\psi))$
            \STATE $\psi \leftarrow$ GradientUpdate$(\psi,d_\psi)$ \label{algo:samplingupdate}
        \ENDIF
    \ENDWHILE
\end{algorithmic}
% \end{multicols}
\end{algorithm}

\subsection{Selection and Sampling Strategies}
\label{sec:sampling_selection}

\subsubsection{Selection}
\label{selection}
The choice of TLPs in meta-learning plays a vital role in influencing the model performance, as we will see in more detail in Section~\ref{sec:results}. Apart from the use of all TLPs across both tasks and languages during training, selecting all languages for a given task  \cite{gu2018meta}  and selecting all tasks for a given language \cite{dou2019investigating} are two other logical choices. We refer to the last two settings as being \texttt{Task-Limited} and \texttt{Lang-Limited}, respectively. 
% Task-language pair (TLP) selection plays a vital role in performance of multi-task meta learning models. We describe our TLPs selection strategies below:
% \begin{itemize}
% \item {\textbf{Task Limited} TL pairs of same task but different language is an obvious choice of auxiliary tasks for each other. It has been widely explored in literature \cite{gu2018meta,dou2019investigating} that using datasets in different languages for the same task admits benefit in meta learning.}
% \item {\textbf{Lang Limited} Similarly, it can argued that using TL pairs of the same language is the next obvious choice for auxiliary tasks.} 
% \end{itemize}

\subsubsection{Heuristic Sampling}
\label{heuristic}

Once the TLPs for meta training (denoted by $\mathcal{D}$) have been selected, we need to sample TLPs from $\mathcal{M}$. We investigate temperature-based heuristic sampling~\cite{Arivazhagan2019MassivelyMN} which defines the probability of any dataset as a function of its size. $P_{\mathcal{D}}(i)$ = $q_{i}^{1/\tau}/\big(\sum_{k = 1}^{n}  q_{k}^{1/\tau}\big)$ where $P_{\mathcal{D}}(i)$ is the probability of the $i^{th}$ TLP to be sampled and $\tau$ is the temperature parameter. $\tau$ = 1 reduces to sampling TLPs proportional to their dataset sizes and $\tau \rightarrow \infty$ reduces to sampling TLPs uniformly.

\subsubsection{Parameterized Sampling}
\label{parametrised}
% TLP sampling probabilities defined in \ref{heuristic} remain constant throughout the meta training. Temperature parameter could be used to adjust the sampling proportion of different TLPs, but weights are still a function of dataset size. To be able to weigh multiple TLPs irrespective of dataset size, we use a scorer network \cite{Wang2020BalancingTF}, which is jointly trained along with the meta model. The scorer network provides weights to TLPs that serves as the probabilities for sampling. In our case, we have a parameter for \vish{Sushil please add a line about scorer network}\\

The sampling strategy defined in Section~\ref{heuristic} remains constant throughout meta training and only depends on dataset sizes. \newcite{Wang2020BalancingTF} proposed a parameterized sampling technique called MultiDDS that builds on Differential Data Selection (DDS)~\cite{differential-rewards} for weighing multiple datasets. The $P_{\mathcal{D}}(i)$ are parameterized using $\psi_{i}$ as $P_\mathcal{D}(i) = e^{\psi_i}/\sum_{j}e^{\psi_j}$ with the initial value of $\psi$ satisfying $P_\mathcal{D}(i) =  q_{i}$. The optimization for $\psi$ and $\theta$ is performed in an 
%boils down to a bi-level 
alternating manner~\cite{colson2007overview}
%wherein,  $\psi$ and $\theta$ are jointly trained in an alternating fashion.

%As we have multiple datasets of different sizes while performing meta-learning, sampling plays an important role in the performance of the model. A good sampling strategy has to account for data imbalance and relationship between tasks. Heuristic based sampling techniques try to combat the issue of data imbalance by oversampling from low resource datasets. One example of such sampling technique is temperature based sampling. However, this technique considers dataset size as the only deciding factor for sampling, and there are other underlying factors which make a task important other than dataset size.

%The MultiDDS \cite{Wang2020BalancingTF} sampling technique tries to find which tasks are more helpful by comparing the first order moments of loss on that task and the objective function on validation set(s) and assigning more weight to task with matching moments. Formally, it tries to solve the following optimization problem:
\vspace{-10pt}
\begin{align}
\psi^{*} &= \underset{\psi}{\operatorname{argmin}} \,\, J(\theta^{*}(\psi), \mathcal{D}_{dev}) \\
\theta^{*}(\psi) &= \underset{\theta}{\operatorname{argmin}} \,\, E_{x,y \sim P(T;\psi)} [l(x, y; \theta)]
\end{align}
$J(\theta,\mathcal{D}_{dev})$ is the objective function which we want to minimize over development set(s). The reward function, $R(x,y;\theta_{t})$, is defined as:
%\begin{align*}
   % \theta_{t} \leftarrow  \theta_{t-1} - \nabla _{\theta} E_{x,y \sim P(x,y;\psi)} [l(x,y;\theta)]
%\end{align*}
\begin{align}
    R(x,\!y;\!\theta_{t}) &\approx \underbrace{\nabla J(\theta_{t},\!\mathcal{D}_{dev})^{T}}_{g_{dev}} \!\cdot\! \underbrace{\nabla_{\theta} l(x,\!y;\!\theta_{t-1})}_{g_{train}} \\
    &\approx cos ( g_{dev}, g_{train}) 
\end{align}
$\psi$'s are updated using the REINFORCE~\cite{williams1992simple} algorithm.
\begin{align}
\psi_{t+1} \!\leftarrow\!  \psi_{t} + R(x,\!y;\!\theta_{t}) \cdot \nabla _{\psi} log (P(x,\!y;\!\psi))
\end{align}

\noindent The Reptile meta-learning algorithm (along with details of the parameterized sampling strategy) is outlined in Algorithm~\ref{algo::reptile}. 

\section{Experimental Setup}
\label{sec:setup}

\subsection{Evaluation Benchmark}
The recently released XTREME dataset~\cite{pmlr-v119-hu20b} is a multilingual multi-task benchmark consisting of classification, structured prediction, QA and retrieval tasks. Each constituent task has associated datasets in multiple languages. The sources of POS and NER datasets are Universal Dependency v2.5 treebank~\cite{nivre2020universal} and WikiAnn~\cite{pan2017cross} respectively, with ground-truth labels available for each language. Large-scale datasets for QA, NLI and PA were originally available only for English. The PAWS-X~\cite{yang2019paws} dataset contains machine-translated training pairs and human-translated evaluation pairs for PA. The authors of XTREME train a custom-built translation system to obtain translated datasets for QA and NLI. For the NLI task, we train using MultiNLI~\cite{N18-1101} and evaluate on XNLI \cite{conneau2018xnli}. For the QA task, SQuAD 1.1~\cite{DBLP:journals/corr/RajpurkarZLL16} was used for training and MLQA~\cite{Lewis2019mlqa} for evaluation. \\

% \noindent\textbf{Preprocessing} - We use the standard preprocessing technique for SQuAD data and BERT tokenizer as defined in \cite{Wolf2019HuggingFacesTS}.\pj{Is this section incomplete?} \ishan{Yes I'll complete it}

\definecolor{green1}{RGB}{24, 106, 59}
\definecolor{green2}{RGB}{29, 131, 72}
\definecolor{green3}{RGB}{248, 196, 113}%{35, 155, 86} 
\definecolor{green4}{RGB}{40, 180, 99}
\definecolor{green5}{RGB}{46, 204, 113}
\definecolor{green6}{RGB}{88, 214, 141}
\definecolor{green7}{RGB}{130, 224, 170}
\definecolor{green8}{RGB}{171, 235, 198}
\definecolor{green9}{RGB}{213, 245, 227}

\definecolor{grad1}{RGB}{181, 247, 205}%{249, 231, 159}
\definecolor{grad2}{RGB}{120, 220, 160}%{248, 196, 113}
\definecolor{grad3}{RGB}{46, 204, 113}%{235, 152, 78}

\begin{table*}[h]
\small
\centering
\captionsetup{font=small}
\begin{tabular*}{\hsize}{@{\extracolsep{\fill}} cccccccc}
\toprule
\multicolumn{2}{c}{Task} & en & hi & es & de & fr & zh \\
\midrule
\multicolumn{2}{c}{Natural Language Inference (NLI)}& \cellcolor{grad2}392K & \cellcolor{red!25} & \cellcolor{grad2}392K & \cellcolor{grad2}392K & \cellcolor{grad2}392K & \cellcolor{red!25} \\
\midrule
\multicolumn{2}{c}{Question Answering (QA)} & \cellcolor{grad2}88.0K & \cellcolor{grad2}82.4K & \cellcolor{grad2}81.8K & \cellcolor{grad2}80.0K & \cellcolor{red!25} &  \cellcolor{red!25} \\ \midrule
\multicolumn{2}{c}{Part Of Speech (POS)}& \cellcolor{grad2}21.2K & \cellcolor{grad2}13.3K & \cellcolor{grad2}28.4K & \cellcolor{grad2}166K & \cellcolor{red!25} & \cellcolor{grad2}7.9K\\
\midrule
\multicolumn{2}{c}{Named Entity Recognition (NER)}& \cellcolor{grad2}20K & \cellcolor{grad2}5K & \cellcolor{grad2}20K & \cellcolor{grad2}20K &  \cellcolor{grad2}20K & \cellcolor{grad2}20K \\
\midrule
\multicolumn{2}{c}{Paraphrase Identification (PA)}& \cellcolor{grad2}49.4K & \cellcolor{red!25} & \cellcolor{grad2}49.4K & \cellcolor{grad2}49.4K & \cellcolor{grad2}49.4K & \cellcolor{grad2}49.4K \\
\bottomrule
\end{tabular*}
\small 
\caption{Dataset matrix showing datasets that are available (green) from the XTREME Benchmark. The number of training instances are also mentioned for each available dataset.} %$|D_{train}^{i}|$}
    \label{tab::datasetdetails}
\end{table*}

\noindent Regarding evaluation metrics, for QA we report F1 scores and for the other four tasks (PA, NLI, POS, NER) we report accuracy scores.

\subsection{Implementation Details}
BERT~\cite{devlinetal2019bert} models yield state-of-the-art performance for many NLP tasks. Since we are dealing with datasets in multiple languages, we build our meta learning models on mBERT \cite{pires2019multilingual,DBLP:journals/corr/abs-1904-09077} base architecture, implemented by \newcite{Wolf2019HuggingFacesTS}, with output layers specific to each task. In our experiments, we use the AdamW~\cite{Loshchilov2017FixingWD} optimizer to make gradient-based updates to the model's parameters using batches from a particular TLP (\hyperref[algo:gradupdate]{Alg. 1, Line 6}). This optimizer is shared across all the TLPs. When performing the meta-step (\hyperref[algo:reptileupdate]{Alg. 1, Line 8}), we use vanilla stochastic gradient descent (SGD) ~\cite{robbins1951stochastic} updates. Similarly, in the case of parameterized sampling the weights are updated (\hyperref[algo:samplingupdate]{Alg. 1, Line 15}) using vanilla SGD. 
%The text files \sushil{corpus} across all languages was tokenized using the BertTokenizer implemented in \newcite{Wolf2019HuggingFacesTS}.
%We chose to implement all the tasks chosen for this work over a BERT architecture. More specifically, since we have datasets in multiple languages, we found mBERT \cite{pires2019multilingual,DBLP:journals/corr/abs-1904-09077} to be an appropriate choice for our base model along with task-specific layers for each task. 

%\subsection{Meta-learning Details}
Meta training involves sampling a set of $m$ tasks, taking $k$ gradient update steps from the initial parameter to arrive at $\theta_{i}^{(k)}$ for task $T_{i}$ and finally updating $\theta$ using the Reptile update rule. For meta-models we fix learning rate = 3e-5 and dropout probability = 0.1 (provided by XTREME for reproduction of baselines). Grid search was performed on $m$ $\in$ \{4, 8, 16\}, $k$ $\in$ \{2, 3, 4, 5\} and $\beta$ $\in$ \{0.1, 0.5, 1.0\} for \texttt{All TLPs} model ($\tau$ = 1). The best setting ($m$ = 8, $k$ = 3, $\beta$ = 1.0) was selected based on validation score (accuracy or F1) averaged over all TLPs. These hyperparameters were kept constant for all further experiments. Each meta-learning model is trained for 5 epochs. We then finetune the meta model individually on each TLP and evaluate the results. Finetuning parameters vary for different task and are mentioned in \hyperref[finetuningSupplementary]{Appendix B}.

%\subsection{Auxiliary TLP Selection with Temperature Sampling}

\subsection{Data Selection and Sampling Strategies}

We experiment with three different configurations for the set of TLPs to be considered during meta-learning: (a) using all tasks for a given language (\texttt{Lang-Limited}) (b) using all languages for a given task (\texttt{Task-Limited}) and (c) using all tasks and all languages (\texttt{All TLPs}). Since the dataset size varies across tasks (as also across languages), we use temperature sampling within each setting for $\tau$ = 1, 2, 5 and $\infty$. (In Table ~\ref{tab:temperatureSamplingResults} of the \hyperref[temperatureSamplingSupplementary]{Appendix C} in the supplementary material, we report results for different choices of TLP selection and different values of the temperature.)

\begin{figure*}
\begin{center}
\captionsetup{font=small}
\includegraphics[width=0.9\textwidth]{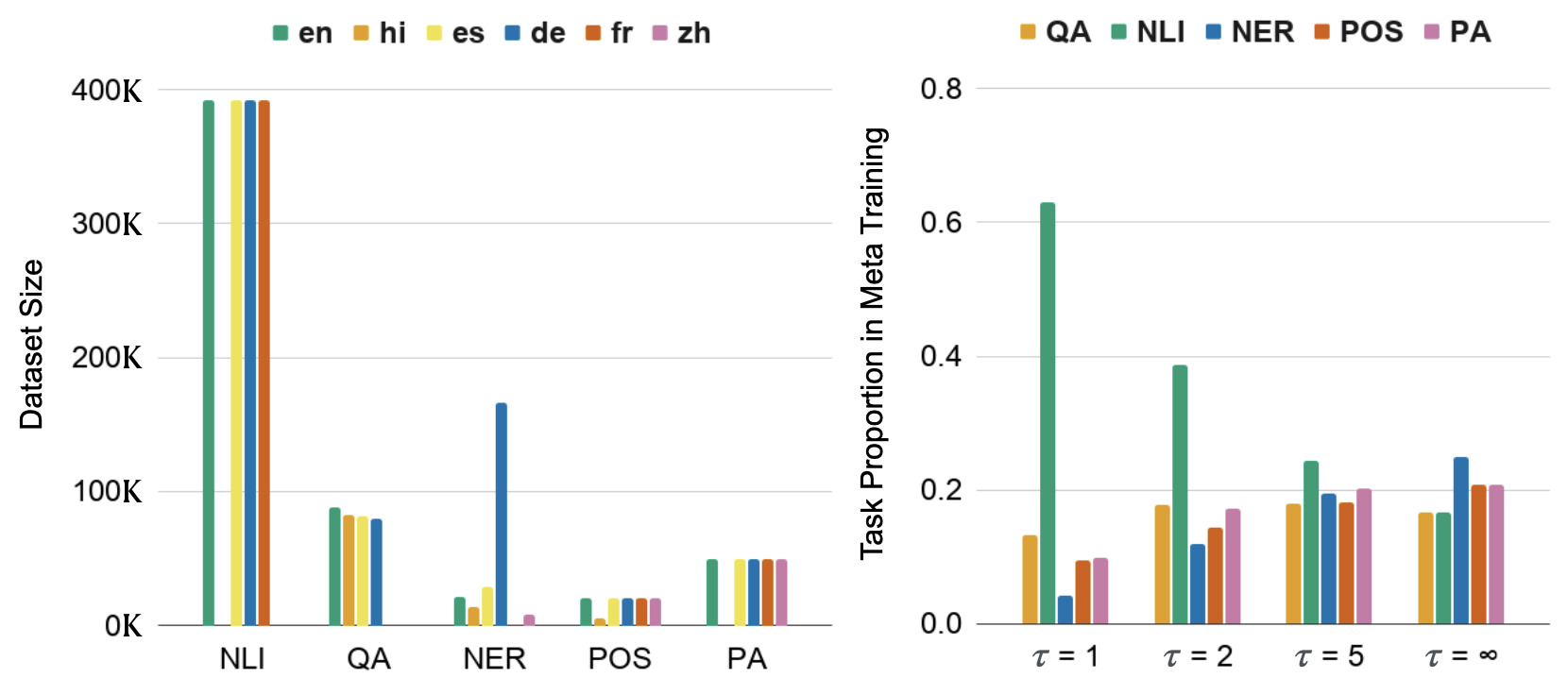}
\captionof{figure}{(a) Size of train dataset by language for each task (b) Proportion of dataset in meta training for different value of $\tau$.}
\label{fig:data}
\end{center}
\end{figure*}

With respect to the~\hyperref[algo::reptile]{\bf Input} in  Algorithm~\ref{algo::reptile},
%Alg. 1, Line 15 ~\ref{algo:samplingupdate}, 
there are two sets of TLPs that need to be selected for parameterized sampling: $\mathcal{D}_{train}$ and $\mathcal{D}_{dev}$. In order to analyse the effect of the choice of task and language, we experiment with the following 4 settings - \\
(a) $\mathcal{D}_{train}$ = \texttt{Lang-Limited}, $\mathcal{D}_{dev}$ = \texttt{Target TLP} \\
(b) $\mathcal{D}_{train}$ = \texttt{Task-Limited}, $\mathcal{D}_{dev}$ = \texttt{Target TLP} \\
(c) $\mathcal{D}_{train}$ = \texttt{All TLPs}, $\mathcal{D}_{dev}$ = \texttt{Lang-Limited} \\
(d) $\mathcal{D}_{train}$ = \texttt{All TLPs}, $\mathcal{D}_{dev}$ = \texttt{Task-Limited}. \\
The models (a), (b) are referred to as \texttt{mDDS}  and (c), (d) are called \texttt{mDDS-Lang} and \texttt{mDDS-Task} respectively. Results for these 4 models are reported in Table~\ref{tab:main} alongside temperature sampling for comparison.

\subsection{Baselines}
Our first baseline system for each TLP uses mBERT-based models trained on data specific to each TLP, which is either available as ground-truth or in a translated form. We follow the same hyperparameter settings as reported in XTREME. We also present three multi-task learning (MTL) baseline systems: task limited (\texttt{Task-Limited}), language limited (\texttt{Lang-Limited}), and the use of all TLPs during training (\texttt{All TLPs MTL}). During MTL training, we concatenate and shuffle the selected datasets. The model is trained for 5 epochs with a learning rate of 5e-5. We refer the reader to \hyperref[baselineSupplementary]{Appendix A} for more training details.
%In order to compare against Multi-task learning we train 3 competitive baselines - Task limited MTL, Lang limited MTL, All TLPs MTL. For training we concat and shuffle the selected datasets (depending on the setting). The model is trained for 5 epochs with a learning rate of 5e-5. Refer to \ref{-} for training details.
%For baselines we use in-language data (available or translated) for each TLP and report mBERT based baseline scores. 

\section{Results and Analysis}
\label{sec:results}

Table~\ref{tab:main} presents all our main results comparing different data selection and sampling strategies used for meta-learning. %We also show mBERT-based baseline systems trained only on data for each TLP and multi-task baseline systems for the three different data selection settings.  
Each column corresponds to a target TLP; the best-performing meta-learned models for each target TLP within each data selection setting have been highlighted in colour. (Light-to-dark gradation reflects improvements in performance.) From Table~\ref{tab:main}, we see that our meta-learned models outperform the baseline systems across all the TLPs corresponding to QA, NLI and PA. (POS and NER also mostly benefit from meta-learning, but the margins of improvement are much smaller compared to the other tasks given the already high baseline scores).

\begin{table*}[!htb]
\scriptsize
\centering
\captionsetup{font=small}
\setlength\tabcolsep{3pt}
\begin{tabular*}{\hsize}{@{\extracolsep{\fill}} @{} *{16}{c} @{} }
\toprule
\multicolumn{2}{c}{\multirow{2}{*}{Model}} & \multirow{2}{*}{SS} & \multicolumn{4}{c}{QA (F1)} & \multicolumn{4}{c}{NLI (Acc.)} & \multicolumn{5}{c}{PA (Acc.)} \\ \cmidrule{4-7} \cmidrule{8-11} \cmidrule{12-16}
& & & en & hi & es & de & en & es & de & fr & en & es & de & fr & zh\\
\midrule
\multicolumn{2}{c}{\texttt{Baselines}} & & \cellcolor{green3}79.94&\cellcolor{green3}59.94&\cellcolor{green3}65.83&\cellcolor{green3}63.17&\cellcolor{green3}81.39&\cellcolor{green3}78.37&\cellcolor{green3}76.82&\cellcolor{green3}77.30& 92.35&89.75&87.45&89.61&\cellcolor{green3}83.32\\
\multicolumn{2}{c}{\texttt{Lang-Limited MTL}} &  & 69.80 & 53.24 & 62.29 & 58.91 & 80.49 & 76.10 & 75.18 & 74.94 & \cellcolor{green3}93.75 & 87.75 & 85.35 & 88.55 & 80.49 \\
\multicolumn{2}{c}{\texttt{Task-Limited MTL}} &  & 74.04 & 57.77 & 64.28 & 61.47 & 80.95 & 78.15 & 75.90 & 77.14 & 93.65 & 86.65 & 86.25 & 86.82 & 81.24 \\
\multicolumn{2}{c}{\texttt{All TLPs MTL}} &  & 63.22 & 42.94 & 54.05 & 51.61 & 80.05 & 76.48 & 74.86 & 76.18 & 93.50 & \cellcolor{green3}90.30 & \cellcolor{green3}88.45 & \cellcolor{green3}89.71 & 82.66 \\

\arrayrulecolor{black!30}\midrule
\multicolumn{2}{c}{\multirow{2}{*}{\texttt{Lang-Limited}}} &\texttt{Temp}&-0.04&-0.24&-0.27&+0.07&\cellcolor{grad1}+0.06&\cellcolor{grad1}+0.39&\cellcolor{grad1}+0.03&-0.70&\cellcolor{grad2}+0.45&\cellcolor{grad2}+0.05&\cellcolor{grad2}+0.35&\cellcolor{grad1}+0.40&\cellcolor{grad1}-0.06\\
&&\texttt{mDDS}&\cellcolor{grad1}+0.07 & \cellcolor{grad1} -0.12 &\cellcolor{grad1} +0.06 & \cellcolor{grad1} +0.14& +0.02&-0.61&-0.80&\cellcolor{grad1}-0.60&-0.25&-0.05&0.00&-0.30&-1.41\\
\arrayrulecolor{black!30}\midrule
\multicolumn{2}{c}{\multirow{2}{*}{\texttt{Task-Limited}}} &\texttt{Temp}&\cellcolor{grad3}+0.55&+0.43&\cellcolor{grad3}+0.50&+0.40&\cellcolor{grad2}+1.65&\cellcolor{grad2}+1.12&+1.25&\cellcolor{grad2}+0.79&+0.20&\cellcolor{grad1}-0.15&-0.55&+0.85&-0.15\\
&&\texttt{mDDS}&+0.21 & \cellcolor{grad3} +0.62 & -0.67 & \cellcolor{grad3} +1.06 & +1.32&+1.10&\cellcolor{grad2}+1.39&+0.48&\cellcolor{grad3}+0.50&-0.65&\cellcolor{grad1}-0.35&\cellcolor{grad3}+1.45&\cellcolor{grad2}+1.06\\
\arrayrulecolor{black!30}\midrule
\multicolumn{2}{c}{\multirow{3}{*}{\texttt{All TLPs}}}& \texttt{Temp}&\cellcolor{grad2}+0.53&+0.47&\cellcolor{grad2}+0.32&+0.47&\cellcolor{grad3}+1.90&\cellcolor{grad3}+1.22&\cellcolor{grad3}+1.45&\cellcolor{grad3}+0.95&\cellcolor{grad1}+0.35&+0.45&+1.20&+1.05&+0.85\\

& &\texttt{mDDS-Lang}& +0.08 & +0.50 & -1.57 & +0.08 & +0.76 & +0.26 & -0.10 & +0.32 & +0.25 & \cellcolor{grad3} +0.85 & +0.75 & +0.75 & +1.11\\
&& \texttt{mDDS-Task}& +0.18 & \cellcolor{grad2} +0.60 &  +0.11 & \cellcolor{grad2} +0.54 & +1.50& +0.90 & +0.72 & +0.72 & +0.10 & +0.80 &\cellcolor{grad3} +1.27 &\cellcolor{grad2} +1.10 &\cellcolor{grad3} +1.16\\
\arrayrulecolor{black}\bottomrule
\end{tabular*}
\begin{tabular*}{\hsize}{@{\extracolsep{\fill}} @{} *{14}{c} @{} }
% \toprule
\multicolumn{2}{c}{\multirow{2}{*}{Model}} & \multirow{2}{*}{SS} & \multicolumn{6}{c}{NER (Acc.)} & \multicolumn{5}{c}{POS (Acc.)} \\ \cmidrule{4-9} \cmidrule{10-14}
& & & en & hi & es & de & fr & zh & en & hi & es & de & zh\\
\midrule
\multicolumn{2}{c}{\texttt{Baselines}}  &&93.23&\cellcolor{green3}95.72&\cellcolor{green3}95.84&\cellcolor{green3}97.32&\cellcolor{green3}95.48&\cellcolor{green3}94.34&\cellcolor{green3}96.15&\cellcolor{green3}93.57&\cellcolor{green3}96.02&\cellcolor{green3}97.37&\cellcolor{green3}92.60\\
\multicolumn{2}{c}{\texttt{Lang-Limited MTL}} && 92.54&92.67&95.14&96.40&94.38&92.97&95.08&92.43&95.19&97.19&89.71\\
\multicolumn{2}{c}{\texttt{Task-Limited MTL}} &&\cellcolor{green3}93.51&93.94&95.77&97.09&95.27&93.72&95.70&93.34&95.73&97.35&92.52\\
\multicolumn{2}{c}{\texttt{All TLPs MTL}} &&92.28&91.95&94.90&96.18&94.38&92.53&94.70&91.89&95.10&97.03&89.92\\
\arrayrulecolor{black!30}\midrule
\multicolumn{2}{c}{\multirow{2}{*}{\texttt{Lang-Limited}}} & \texttt{Temp} &\cellcolor{grad2}+0.60&\cellcolor{grad3}+0.06&\cellcolor{grad2}+0.09&\cellcolor{grad3}+0.24&\cellcolor{grad2}-0.09&\cellcolor{grad2}-0.47&\cellcolor{grad3}-0.06&\cellcolor{grad3}-0.01&\cellcolor{grad3}+0.10&\cellcolor{grad2}+0.04&\cellcolor{grad1}-0.17\\
& & \texttt{mDDS} &-0.21&-0.85&-0.20&-0.10&-0.57&-0.55&-0.27&-0.02&-0.19&-0.06&-0.37\\
\arrayrulecolor{black!30}\midrule

\multicolumn{2}{c}{\multirow{2}{*}{\texttt{Task-Limited}}} & \texttt{Temp} &\cellcolor{grad3}+0.79&\cellcolor{grad1}-0.46&\cellcolor{grad1}0.00&\cellcolor{grad1}-0.07&\cellcolor{grad1}-0.18&\cellcolor{grad1}-0.51&\cellcolor{grad1}-0.22&-0.05&\cellcolor{grad1}-0.21&+0.02&\cellcolor{grad3}-0.09\\
& & \texttt{mDDS} &-0.10&-1.61&\cellcolor{grad1}0.00&-0.16&-0.33&-0.69&-0.38&\cellcolor{grad2}-0.02&-0.22&\cellcolor{grad3}+0.05&-0.12\\
\arrayrulecolor{black!30}\midrule
\multicolumn{2}{c}{\multirow{3}{*}{\texttt{All TLPs}}} & \texttt{Temp} &\cellcolor{grad1}-0.15&-0.70&\cellcolor{grad3}+0.13&\cellcolor{grad2}0.00&-0.16&\cellcolor{grad3}-0.39&-0.22&\cellcolor{grad1}-0.09&-0.21&\cellcolor{grad1}+0.03&-0.16\\
&& \texttt{mDDS-Lang} &-0.16 & \cellcolor{grad2}-0.09 & +0.11 & -0.08 & -0.14 & -0.65 & \cellcolor{grad2} -0.21 & -0.10 & \cellcolor{grad2} -0.11 &\cellcolor{grad1} +0.03 & -0.17\\
&& \texttt{mDDS-Task} &-0.27 & -0.42 & +0.08 & -0.14 & \cellcolor{grad3} -0.07 & -0.58 & -0.22 & -0.14 & -0.19 & +0.02 & \cellcolor{grad3}-0.09\\
\arrayrulecolor{black}\bottomrule
\end{tabular*}
\caption{Main results comparing different data selection and sampling strategies. Sampling strategy, SS=\texttt{Temp} refers to the temperature-based sampling strategy and SS=\texttt{mDDS} refers to the multiDDS-based sampling strategy. \texttt{mDDS-Task} and \texttt{mDDS-Lang} refer to the use of a development set for multiDDS that contains all languages for a task and all tasks for a language, respectively. The best result among Baseline and three MTL models is highlighted using orange. For each column we present the difference (positive or negative) of the meta models from the best baseline (highlighted in orange) of that column}
    \label{tab:main}
\end{table*}

\paragraph{\texttt{Task-Limited} vs \texttt{Lang-Limited} models.} For QA and NLI, we observe that the \texttt{Task-Limited} models are {\em always better} than the \texttt{Lang-Limited} models. This is in line with our intuition that tasks like QA and NLI (which require deeper semantic representations) will benefit more by using data from different languages for the same task. We see the opposite seems to hold for POS and NER where the \texttt{Lang-Limited} models are {\em almost always better} than the \texttt{Task-Limited} models. With POS and NER being relatively shallower tasks, it makes sense that they benefit more from language-specific training that relies on token embeddings shared across tasks. 

\paragraph{Investigating Sampling Strategies.} 
In Table~\ref{tab:main}, all the scores shown for the \texttt{Temp} sampling strategy are the best scores across four different values of $T$, $T = 1,2,5,\infty$. (The complete table is available in \hyperref[tab:temperatureSamplingResults]{Appendix C} in the supplementary material.) 
\begin{figure*}[h]
\begin{center}
	\captionsetup{font=small}
	\includegraphics[width=1.0\textwidth]{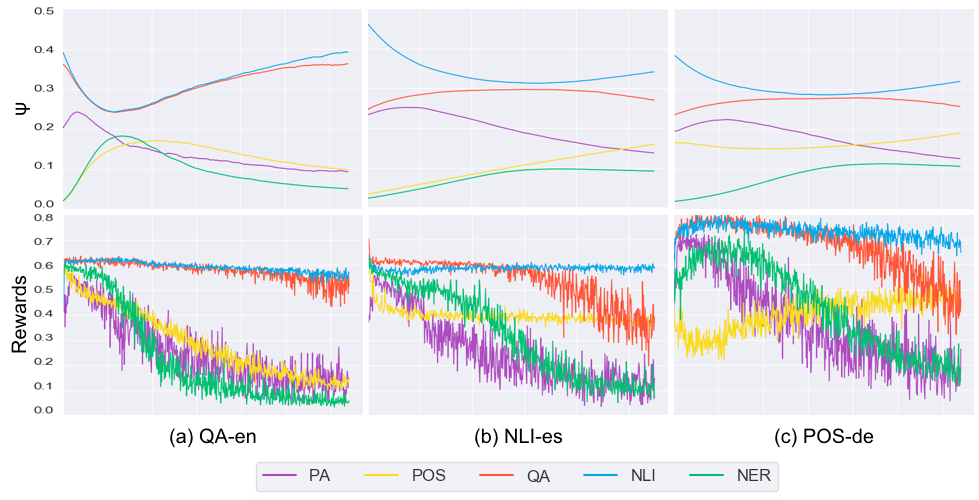}
  	\captionof{figure}{Evolution of $\psi$s and rewards as a function of training time for three \texttt{Lang-Limited} tasks evaluated on (a) QA-en (b) NLI-es and (c) POS-de.}
  	\label{fig:mdds}
\end{center}
\end{figure*}
We also present comparisons with the \texttt{mDDS}, \texttt{mDDS-Lang} and \texttt{mDDS-Task} sampling strategies enforced within the \texttt{Lang-Limited}, \texttt{Task-Limited} and \texttt{All TLPs} models, respectively. For POS and NER, our best meta-learned models are mostly \texttt{Lang-Limited} with \texttt{Temp} sampling. It is intuitive that for these shallower tasks, \texttt{mDDS} does not offer any benefits from allowing to sample instances from other tasks. 
%Hence, we focus on the remaining three tasks: QA, NLI and PA. 

%Figure~\ref{fig:mdds} shows how $\psi$'s and the rewards change over time for three \texttt{Lang-limited} tasks evaluated on the target TLPs QA-en, NLI-es and PA-de. 

\begin{table*}[!htb]
\scriptsize
\centering
\captionsetup{font=small}
\setlength\tabcolsep{3pt}
\begin{tabular*}{\hsize}{@{\extracolsep{\fill}} @{} *{17}{c} @{} }
% \toprule
\toprule
\multicolumn{2}{c}{\multirow{2}{*}{Model}}  & \multicolumn{8}{c}{NER (Acc.)} & \multicolumn{7}{c}{POS (Acc.)} \\ \cmidrule{3-10} \cmidrule{11-17}
& & bn & et & fi & ja & mr & ta & te & ur & et & fi & ja & mr & ta & te & ur\\
\midrule
\multicolumn{2}{c}{\texttt{Task-Limited MTL}}&\cellcolor{green3}81.80&\cellcolor{green3}93.98&\cellcolor{green3}94.47&\cellcolor{green3}81.03&\cellcolor{green3}90.63&\cellcolor{green3}83.46&\cellcolor{green3}87.67&\cellcolor{green3}69.25&\cellcolor{green3}85.21&\cellcolor{green3}83.98&\cellcolor{green3}58.42&\cellcolor{green3}72.56&\cellcolor{green3}73.88&\cellcolor{green3}79.15&\cellcolor{green3}86.08\\
\multicolumn{2}{c}{\texttt{All TLPs MTL}}&77.49&90.35&92.65&77.80&81.19&81.21&86.17&64.27&69.63&73.50&57.24&68.80&70.52&72.41&81.59\\
\arrayrulecolor{black!30}\midrule
\multicolumn{2}{c}{\texttt{Task-Limited}} & \cellcolor{grad2}+1.91 & \cellcolor{grad2}+0.63 & \cellcolor{grad2} +0.16 & \cellcolor{grad2}+0.35 & \cellcolor{grad2}-0.67 & \cellcolor{grad2}+1.34 & \cellcolor{grad2}+0.63 & +2.14 & \cellcolor{grad2} +2.94 & \cellcolor{grad2} +2.15 & \cellcolor{grad2} +0.83 & \cellcolor{grad2} +8.64 & \cellcolor{grad2} +2.34 & \cellcolor{grad2} +2.82 & -0.30\\

\multicolumn{2}{c}{\texttt{All TLPs}} & +0.62 & +0.35 & -0.11 & +0.19 & -0.92 & +1.25 & +0.43 & \cellcolor{grad2} +9.10 & +2.56 & +2.01 & -1.42 & +8.27 & +1.24 & +2.51 & \cellcolor{grad2}-0.16\\ 

\multicolumn{2}{c}{\texttt{All TLPs mDDS-Task}} & -0.83 & +0.09 & -0.20 & -1.34 & -1.87 & +0.49 & +0.05 & +3.62 & +1.91 & +1.08 & -1.74 & \cellcolor{grad2} +8.64 & +1.24 & +1.88 & -0.72\\
\arrayrulecolor{black}\bottomrule
\end{tabular*}
\begin{tabular*}{\hsize}{@{\extracolsep{\fill}} @{} *{15}{c} @{} }
\multicolumn{2}{c}{\multirow{2}{*}{Model}} & \multicolumn{2}{c}{QA (F1)} & \multicolumn{9}{c}{NLI (Acc.)} & \multicolumn{2}{c}{PA (Acc.)} \\ \cmidrule{3-4} \cmidrule{5-13} \cmidrule{14-15}
& & ar & vi & ar & bg & el & ru & sw & th & tr & ur & vi & ja & ko\\
\midrule
\multicolumn{2}{c}{\texttt{Task-Limited MTL}}&32.25&44.35&62.88&67.47&66.09&67.85&43.61&43.16&\cellcolor{green3}57.79&\cellcolor{green3}57.03&69.45&\cellcolor{green3}78.23&\cellcolor{green3}74.85\\
\multicolumn{2}{c}{\texttt{All TLPs MTL}}&\cellcolor{green3}40.14&\cellcolor{green3}54.08&\cellcolor{green3}64.54&\cellcolor{green3}67.99&\cellcolor{green3}66.25&\cellcolor{green3}70.05&\cellcolor{green3}43.89&\cellcolor{green3}45.72&56.73&56.93&\cellcolor{green3}72.02&77.61&73.49\\
\arrayrulecolor{black!30}\midrule
\multicolumn{2}{c}{\texttt{Task-Limited}} & \cellcolor{grad2} +8.14 &\cellcolor{grad2} +6.63 & +4.35 & \cellcolor{grad2}+5.15 & +4.62 & +2.72 & +8.51 & \cellcolor{grad2} +14.42 & +6.79 & +5.27 & \cellcolor{grad2} +1.3 & +0.21 & +1.81\\

\multicolumn{2}{c}{\texttt{All TLPs}} & +5.24 & +3.62 & \cellcolor{grad2}+4.41 & +4.73 & \cellcolor{grad2} +4.79 & \cellcolor{grad2}+2.94 & \cellcolor{grad2} +11.44 & +13.04 & +7.05 & \cellcolor{grad2} +5.67 & +1.24 & \cellcolor{grad2} +3.07 & +4.57\\ 
\multicolumn{2}{c}{\texttt{All TLPs mDDS-Task}} & +6.89 & +6.29 & +3.19 & +4.33 & +4.09 & +2.38 & +8.71 & +13.16 & \cellcolor{grad2} +7.09 & +4.41 & +1.04 & +2.81 & \cellcolor{grad2} +4.92\\
\arrayrulecolor{black}\bottomrule
\end{tabular*}
\caption{Results comparing Zero-shot evaluations for several external languages with competitive MTL baselines. The best MTL model is highlighted using orange. Rows for meta models show the difference (positive or negative) of the meta model result from the best MTL setting (orange) for that column}
    \label{tab:zero_shot_results}
\end{table*}

To better understand the effects of \texttt{mDDS} sampling, Figure~\ref{fig:mdds} shows plots of the rewards and sampling probabilities $\psi$'s computed as a function of training time for two deeper tasks - QA-en and NLI-es along with a shallower task - POS-de. We note that initially all the TLPs in any \texttt{mDDS} setting would start with similar rewards, thus lending $\psi$'s to converge towards the $T=\infty$ state. 
We highlight the following three observations:
\begin{itemize}
    \item We find that the \texttt{mDDS} strategy does not help NLI at all. This is because the NLI task occupies the largest proportion across tasks at the start, as shown in Figure~\ref{fig:data}, and the proportion of NLI decreases substantially over time (since all tasks start with similar rewards at the beginning of meta training). Thus, for tasks that are over-represented in the meta-learning phase, temperature-based sampling is likely to be sufficient. 
    \item We observe that the rewards for both QA and NLI are consistently high, irrespective of the target TLP. This suggests that both QA and NLI are information-rich tasks and could benefit other tasks in meta-learning. This is also apparent from the accuracies for PA in Table~\ref{tab:main}, where all the best meta-learned models employ \texttt{mDDS} sampling.
    \item From the sampling probabilities for QA-en, we see that both QA and NLI are given almost equal weightage. However, from the F1 scores in Table~\ref{tab:main}, the best numbers for QA are in the \texttt{Task-Limited} setting which suggests that QA does not benefit from any other task. One explanation for this could be that the sequence length of inputs for NLI is 128 while the inputs for QA are of length 384, thus allowing lesser room for QA to be benefited by NLI.
\end{itemize}

%\noindent Initially all the TLPs in any MultiDDS setting would have similar rewards leading to $\psi$'s converging towards T = $\infty$. The PA task is very similar to NLI (both being part of GLUE \cite{wang2019glue}) but we observe that the reward of PA task when target is NLI (or vice versa) is low. TARGET-TASK models have all TLPs as part of their train set but only TLPs limited to one task in their development set. The temperature models perform better for NLI compared to TARGET-TASK because of two reasons - firstly the proportion of NLI decreases substantially in the start because of similar rewards for all task at the start of meta training (NLI has the highest proportion initially) and secondly because maximizing performance on NLI in not guaranteed to converge to a better set of parameters for finetuning.

\paragraph{Zero-shot Evaluations.} 
Zero-shot evaluation is performed on languages that were not part of the training (henceforth, we refer to  them as external languages). In the case of QA, NLI and PA we select all external language for which datasets were available in XTREME. For NER and POS, the number of external languages is close to 35 so we choose a subset of these to report the results. For evaluation, we compare models that are agnostic to the target language during meta training  (\texttt{Task-Limited}, \texttt{All TLPs} and \texttt{All TLPs mDDS-Task}). Since \texttt{Lang-Limited MTL} is language specific and does not offer a competitive baseline when applied to an external language, we  compare against  \texttt{Task-Limited MTL} and \texttt{All TLPs MTL} that are more competitive.

An interesting observation from the zero shot results in Table~\ref{tab:zero_shot_results} is that for every external language, on the `shallower' NER and POS tasks, the \texttt{Task-Limited} variant of meta-learning performs better than both the variants of MTL, {\em viz.}, \texttt{Task-Limited MTL} and \texttt{All TLPs MTL}. In contrast, the `deeper' tasks, {\em viz.}, QA, NLI and PA benefit more from the use of meta-learning using  \texttt{All TLPs} setting, presumably because, as argued earlier, the deeper tasks tend to help each other more.

\section{Conclusion}

We present effective use of meta-learning for capturing task and language interactions in multi-task, multi-lingual settings. The effective use involves appropriate strategies for sampling tasks and languages as well as rough knowledge of the level of abstraction (deep vs. shallow representation) of that task.  We present experiments on the XTREME multilingual benchmark dataset using five tasks and six languages. Our meta-learned model shows clear performance improvements over competitive baseline models. We observe that deeper tasks  consistently benefit from meta-learning. Furthermore, shallower tasks benefit from deeper tasks when meta-learning is restricted to a single language. Finally, zero-shot evaluations for several external languages demonstrate the benefit of using meta-learning over two multi-task baselines while also reinforcing the linguistic insight that tasks requiring deeper representations tend to collaborate better.
% In this work, we looked at using meta-learning in a multitask multilingual setting. We found that leveraging information from other tasks and languages can help us obtain a more generalised model. We also compared heuristic based sampling with parameterised sampling and find that the ideal choice of sampling technique depends on the target task in our setting.

\section*{Acknowledgements}
We thank anonymous reviewers for providing constructive feedback. We are grateful to IBM Research, India (specifically the IBM AI Horizon Networks - IIT Bombay initiative) for their support and sponsorship.

%\bibliographystyle{acl_natbib}
%\bibliography{anthology,eacl2021}

\clearpage
\section*{Appendices}

\subsection*{Appendix A: Baseline Training Details}
\label{baselineSupplementary}

For QA learning rate is 3e-5 and sequence length is 384 and the model is trained for 2 epochs. For PA, NLI, POS and NER the learning rate is 2e-5 and sequence length is 128. NLI and PA models are trained for 5 epochs while POS and NER models are trained for 10 epochs. The choice of hyperparameters was kept constant across different languages for the same task. 

\subsection*{Appendix B: Finetuning Details}
For finetuning we kept the same number of epochs as the baseline of that task i.e 2 epochs for QA, 10 epochs for POS and NER, 5 epochs for NLI and PA. For QA we finetune with learning rate 3e-5 and 3e-6 and POS/NER we finetune with learning rate 2e-5 and 2e-6 and select the better of the two model. For PA and NLI the results for learning rate 2e-5 were consistently worse compared to 2e-6 so we just use lr = 2e-6 for PA and NLI. 

\label{finetuningSupplementary}
\subsection*{Appendix C: Temperature Sampling}
\label{temperatureSamplingSupplementary}

\begin{table*}[h]
\scriptsize
\centering
\captionsetup{font=small}
\setlength\tabcolsep{3pt}
\begin{tabular*}{\hsize}{@{\extracolsep{\fill}} @{} *{16}{c} @{} }
\toprule
\multicolumn{2}{c}{\multirow{2}{*}{Model}} & \multirow{2}{*}{T} & \multicolumn{4}{c}{QA (F1)} & \multicolumn{4}{c}{NLI (Acc.)} & \multicolumn{5}{c}{PA (Acc.)} \\ \cmidrule{4-7} \cmidrule{8-11} \cmidrule{12-16}
& & & en & hi & es & de & en & es & de & fr & en & es & de & fr & zh\\
\midrule
\multicolumn{2}{c}{\texttt{Baselines}} & & \cellcolor{green3!50}79.94&\cellcolor{green3!50}59.94&\cellcolor{green3!50}65.83&\cellcolor{green3!50}63.17&\cellcolor{green3!50}81.39&\cellcolor{green3!50}78.37&\cellcolor{green3!50}76.82&\cellcolor{green3!50}77.30& 92.35&89.75&87.45&89.61&\cellcolor{green3!50}83.32\\
\multicolumn{2}{c}{\texttt{Lang-Limited MTL}} &  & 69.80 & 53.24 & 62.29 & 58.91 & 80.49 & 76.10 & 75.18 & 74.94 & \cellcolor{green3!50}93.75 & 87.75 & 85.35 & 88.55 & 80.49 \\
\multicolumn{2}{c}{\texttt{Task-Limited MTL}} &  & 74.04 & 57.77 & 64.28 & 61.47 & 80.95 & 78.15 & 75.90 & 77.14 & 93.65 & 86.65 & 86.25 & 86.82 & 81.24 \\
\multicolumn{2}{c}{\texttt{All TLPs MTL}} &  & 63.22 & 42.94 & 54.05 & 51.61 & 80.05 & 76.48 & 74.86 & 76.18 & 93.50 & \cellcolor{green3!50}90.30 & \cellcolor{green3!50}88.45 & \cellcolor{green3!50}89.71 & 82.66 \\
\arrayrulecolor{black!30}\midrule
\multicolumn{2}{c}{\multirow{4}{*}{\texttt{Lang-Limited}}} 
& T = 1 &79.49&59.42&64.67&63.04&81.13&\cellcolor{grad1}78.76&76.23&76.51&93.85&89.15&87.83&89.63&82.56\\
& & T = 2 &78.81&59.68&65.10&\cellcolor{grad1}63.24&80.87&77.56&\cellcolor{grad1}76.85&\cellcolor{grad1}76.60&93.85&90.15&87.70&89.41&83.10\\
& & T = 5 &\cellcolor{grad1}79.90&58.74&\cellcolor{grad1}65.56&62.12&81.19&78.17&76.10&76.56&93.65&\cellcolor{grad2}90.35&88.60&\cellcolor{grad1}90.11&83.20\\
& & T = $\infty$ &79.71&\cellcolor{grad1}59.70&65.29&62.89&\cellcolor{grad1}81.45&78.45&76.74&76.46&\cellcolor{grad3}94.20&89.65&\cellcolor{grad2}88.80&89.56&\cellcolor{grad2}83.26\\

\arrayrulecolor{black!30}\midrule
\multicolumn{2}{c}{\multirow{4}{*}{\texttt{Task-Limited}}} 
& T = 1 &80.30&\cellcolor{grad2}60.37&66.32&\cellcolor{grad2}63.57&82.91&\cellcolor{grad2}79.49&77.96&78.02&\cellcolor{grad1}93.95&\cellcolor{grad1}90.15&87.50&\cellcolor{grad2}90.56&82.66\\
& & T = 2 &79.95&59.94&\cellcolor{grad3}66.33&63.50&83.03&79.41&77.94&78.08&93.05&89.85&\cellcolor{grad1}87.90&89.66&\cellcolor{grad1}83.17\\
& & T = 5 &\cellcolor{grad3}80.49&60.17&65.94&62.74&82.75&79.33&77.98&78.00&93.90&89.80&87.65&90.21&83.12\\
& & T = $\infty$ &79.77&59.86&66.01&62.96&\cellcolor{grad2}83.03&79.39&\cellcolor{grad2}78.07&\cellcolor{grad2}78.09&93.60&89.75&87.75&89.61&82.42\\
\arrayrulecolor{black!30}\midrule
\multicolumn{2}{c}{\multirow{4}{*}{\texttt{All TLPs}}} 
& T = 1 &80.20&59.89&66.10&\cellcolor{grad3}63.64&\cellcolor{grad3}83.29&\cellcolor{grad3}79.59&77.84&78.19&93.90&89.95&88.70&90.41&83.57\\
& & T = 2 &\cellcolor{grad2}80.47&\cellcolor{grad3}60.41&66.04&63.56&82.71&78.83&77.96&78.04&93.50&\cellcolor{grad3}90.75&\cellcolor{grad3}89.65&90.71&84.02\\
& & T = 5 &80.01&59.38&\cellcolor{grad2}66.15&63.53&83.19&79.51&78.10&78.21&\cellcolor{grad2}94.10&90.05&88.70&90.26&\cellcolor{grad3}84.17\\
& & T = $\infty$ &80.27&59.82&64.41&63.08&83.27&79.43&\cellcolor{grad3}78.27&\cellcolor{grad3}78.25&94.05&\cellcolor{grad3}90.75&88.70&\cellcolor{grad3}90.76&83.42\\

\arrayrulecolor{black}\bottomrule
\end{tabular*}
\begin{tabular*}{\hsize}{@{\extracolsep{\fill}} @{} *{14}{c} @{} }
% \toprule
\multicolumn{2}{c}{\multirow{2}{*}{Model}} & \multirow{2}{*}{T} & \multicolumn{6}{c}{NER (Acc.)} & \multicolumn{5}{c}{POS (Acc.)} \\ \cmidrule{4-9} \cmidrule{10-14}
& & & en & hi & es & de & fr & zh & en & hi & es & de & zh\\
\midrule
\multicolumn{2}{c}{\texttt{Baselines}}  &&93.23&\cellcolor{green3!50}95.72&\cellcolor{green3!50}95.84&\cellcolor{green3!50}97.32&\cellcolor{green3!50}95.48&\cellcolor{green3!50}94.34&\cellcolor{green3!50}96.15&\cellcolor{green3!50}93.57&\cellcolor{green3!50}96.02&\cellcolor{green3!50}97.37&\cellcolor{green3!50}92.60\\
\multicolumn{2}{c}{\texttt{Lang-Limited MTL}} && 92.54&92.67&95.14&96.40&94.38&92.97&95.08&92.43&95.19&97.19&89.71\\
\multicolumn{2}{c}{\texttt{Task-Limited MTL}} &&\cellcolor{green3!50}93.51&93.94&95.77&97.09&95.27&93.72&95.70&93.34&95.73&97.35&92.52\\
\multicolumn{2}{c}{\texttt{All TLPs MTL}} &&92.28&91.95&94.90&96.18&94.38&92.53&94.70&91.89&95.10&97.03&89.92\\
\arrayrulecolor{black!30}\midrule
\multicolumn{2}{c}{\multirow{4}{*}{\texttt{Lang-Limited}}} & T = 1 &93.14&95.36&95.40&97.21&\cellcolor{grad3}95.39&93.63&95.96&93.33&95.81&97.32&92.32\\
& & T = 2 &93.24&94.76&95.80&\cellcolor{grad3}97.56&95.07&93.53&95.87&93.53&95.93&97.39&92.40\\
& & T = 5 &94.03&\cellcolor{grad3}95.78&\cellcolor{grad2}95.93&97.24&94.99&93.60&\cellcolor{grad3}96.09&\cellcolor{grad3}93.56&95.85&97.33&\cellcolor{grad1}92.43\\
& & T = $\infty$ &\cellcolor{grad2}94.11&95.40&95.75&96.89&95.35&\cellcolor{grad2}93.87&95.99&93.28&\cellcolor{grad3}96.12&\cellcolor{grad3}97.41&92.35\\
\arrayrulecolor{black!30}\midrule
\multicolumn{2}{c}{\multirow{4}{*}{\texttt{Task-Limited}}}& T = 1 &\cellcolor{grad3}94.30&\cellcolor{grad2}95.26&95.82&\cellcolor{grad1}97.25&95.26&93.62&\cellcolor{grad1}95.93&93.36&\cellcolor{grad1}95.81&97.31&92.38\\
& & T = 2 &93.30&94.92&95.82&97.07&\cellcolor{grad1}95.30&93.63&95.84&\cellcolor{grad2}93.52&95.78&97.31&92.38\\
& & T = 5 &93.29&95.02&95.73&96.98&95.19&93.56&95.92&93.34&95.75&\cellcolor{grad1}97.39&92.43\\
& & T = $\infty$ &93.37&94.70&\cellcolor{grad1}95.84&96.95&95.20&\cellcolor{grad1}93.83&95.77&93.33&95.76&97.33&\cellcolor{grad3}92.51\\
\arrayrulecolor{black!30}\midrule
\multicolumn{2}{c}{\multirow{4}{*}{\texttt{All TLPs}}} & T = 1 &93.14&93.63&95.91&97.30&\cellcolor{grad2}95.32&93.53&95.90&93.35&95.76&97.36&92.43\\
& & T = 2 &93.35&\cellcolor{grad1}95.02&95.78&97.30&95.29&93.58&95.92&\cellcolor{grad1}93.48&\cellcolor{grad1}95.81&97.39&\cellcolor{grad2}92.44\\
& & T = 5 &\cellcolor{grad2}\cellcolor{grad1}93.36&94.51&95.93&97.26&95.28&\cellcolor{grad3}93.95&95.92&93.35&95.78&\cellcolor{grad2}97.40&92.42\\
& & T = $\infty$ &93.35&94.95&\cellcolor{grad3}95.97&\cellcolor{grad2}97.32&95.28&93.63&\cellcolor{grad1}95.93&93.31&95.80&97.30&92.43\\
\arrayrulecolor{black}\bottomrule
\end{tabular*}
\caption{Detailed results of temperature based heuristic sampling for different selections settings. The best result among Baseline and three MTL models is highlighted using orange. For each column we present the difference (positive or negative) of the meta models from the best baseline (highlighted in orange) of that column}
    \label{tab:temperatureSamplingResults}
\end{table*}

\end{document}